\documentclass[letterpaper, 10 pt, conference]{IEEEconf} 

\IEEEoverridecommandlockouts

\usepackage{cite}
\usepackage{amsmath,amssymb,amsfonts}
\usepackage{bm}
\usepackage{algorithmic}
\usepackage{graphicx}
\usepackage{textcomp}
\usepackage{xcolor}
\usepackage{balance}
\usepackage[hidelinks]{hyperref}
\usepackage{lipsum}
\usepackage{breqn}
\usepackage{booktabs}
\usepackage{siunitx}
\usepackage{}
\usepackage{import}
\usepackage{standalone}
\usepackage{tikz}
\usetikzlibrary{calc,patterns,decorations.pathmorphing,decorations.markings,positioning,backgrounds,arrows,shapes,fit,matrix,spy,shapes.geometric}
\usepackage{pgfplots}
\pgfplotsset{width=10cm,compat=1.9}

\renewcommand{\vec}[1]{\boldsymbol{#1}}
\newcommand{\mat}[1]{\mathbf{#1}}
\newcommand{\quat}{\vec q}

\newcommand{\xtilde}{\Tilde{\vec x}}
\newcommand{\dxtilde}{\dot{\xtilde}}
\newcommand{\ptilde}{\Tilde{\vec p}}
\newcommand{\dptilde}{\dot{\ptilde}}
\newcommand{\qtilde}{\Tilde{\quat}}
\newcommand{\dqtilde}{\dot{\qtilde}}
\newcommand{\diag}[1]{
    \textrm{diag}
    \left\{#1\right\}
    }
\newcommand{\blkdiag}[1]{
    \textrm{blkdiag}
    \left\{#1\right\}
    }

\begin{document}

\title{\LARGE \bf A Passivity-Based Variable Impedance Controller for Incremental Learning of Periodic Interactive Tasks
\thanks{We acknowledge the support of the MUR PNRR project FAIR - Future AI Research (PE00000013), European Union projects INVERSE (grant agreement no. 101136067) and MAGICIAN (grant agreement no. 101120731)}}

\author{Matteo Dalle Vedove$^{1,2}$, Edoardo Lamon$^{3,4}$, Daniele Fontanelli$^{1}$, Luigi Palopoli$^{3}$, and Matteo Saveriano$^{1}$
\thanks{$^{1}$Department of Industrial Engineering, Universit\`a di Trento, Trento, Italy. \tt\small matteo.dallevedove@unitn.it}
\thanks{$^{2}$DRIM, Ph.D. of national interest in Robotics and Intelligent Machines.}
\thanks{$^{3}$Department of Information Engineering and Computer Science, Universit\`a di Trento, Trento, Italy. }
\thanks{$^{4}$Human-Robot Interfaces and Interaction, Istituto Italiano di Tecnologia, Genoa, Italy.}
}

\maketitle

\thispagestyle{empty}
\pagestyle{empty}

\begin{abstract}
In intelligent manufacturing, robots are asked to dynamically adapt their behaviours without reducing productivity. Human teaching, where an operator physically interacts with the robot to demonstrate a new task, is a promising strategy to quickly and intuitively reconfigure the production line. However, physical guidance during task execution poses challenges in terms of both operator safety and system usability. In this paper, we solve this issue by designing a variable impedance control strategy that regulates the interaction with the environment and the physical demonstrations, explicitly preventing at the same time passivity violations. We derive constraints to limit not only the exchanged energy with the environment but also the exchanged power, resulting in smoother interactions. By monitoring the energy flow between the robot and the environment, we are able to distinguish between disturbances (to be rejected) and physical guidance (to be accomplished), enabling smooth and controlled transitions from teaching to execution and vice versa. The effectiveness of the proposed approach is validated in wiping tasks with a real robotic manipulator.    
\end{abstract}

\section{Introduction}
The introduction of collaborative robotic arms in manufacturing processes brought several benefits in terms of safety, flexibility, and ease of use. 
The limitations in actuation torque contribute to intrinsic safety by constraining the robot physical capabilities. Moreover, their control architecture can incorporate compliant behaviours by dynamically regulating the robot impedance of the robot based on the task requirements and safety protocols, enabling safe interaction with unstructured environments.
Second, it is possible to quickly teach new skills from human demonstrations to the robot intuitively. The versatility in the execution of a variety of tasks, combined with the need of adapting the production process to new products, makes collaborative robot a flexible tool for reconfigurable manufacturing systems.

In this context, Programming by Demonstration (PbD) represent one of the noteworthy methods to transfer physical skills from human demonstrations,
making robot programming faster, easier and more accessible to people without prior robotics expertise, such as factory workers.
To learn from multiple demonstrated trajectories, PbD requires a strategy to parameterise and encode the taught demonstration. 
Examples of the most used are (task-parameterized) Gaussian Mixture Model (GMM)~\cite{calinon2016tutorial}, Dynamical Movement Primitives (DMPs)~\cite{Ijspeert2013, saveriano2023dynamic}, Probabilistic Movement Primitives (ProMP)~\cite{paraschos2013probabilistic}, and Kernelized Movement Primitives (KMP)~\cite{huang2019kernelized}.
The demonstrated trajectories could be obtained from visual observations or from kinesthetic teaching~\cite{billard2016learning}. In the context of collaborative robots, whose design envisions physical interactions with humans, the latter has proven to be more effective, since it does not require any retargeting from the teacher to the robot.
While at the kinematic level human demonstrations have proven a high level of reliability and precision in repetitions, learning contact forces through kinesthetic teaching is still complex due to the high variability of the human ability to accurately exert the same amount of force~\cite{rozo2016learning}. For this reason, in the case of physically interactive tasks, PbD needs to be coupled with adaptive control schemes that are able to adjust the interaction forces with the environment and the teacher. This regulation is particularly complex in the case of incremental or continual learning applications, where the learning phase is interleaved on-the-fly with the execution phase. 

In this direction, variable impedance strategies allow the robot to interact safely with the environment, avoiding the application of high forces that could harm the human or damage the robot or the surroundings, especially in case of unexpected perturbation of the system, while at the same time ensuring accurate trajectory tracking according to the task requirements~\cite{saveriano2023dynamic}. 
This ability, in addition, if coupled with a system capable of distinguishing the desired physical interactions from the unexpected ones, enables the robot to behave differently in teaching (being more compliant) and execution phases (being stiffer).

Several examples of PbD with variable interactive controllers are available in literature~\cite{gams2016adaptation,Kastritsi2018Progressive, Dimes2020Progressive}.
For instance, in~\cite{Dimes2020Progressive} authors exploit two DMPs to learn the Cartesian motion and the interaction forces, and replicate it with a hybrid force-position controller. The teaching phase is achieved through gravity compensation obtained by reducing the controller stiffness according to the Cartesian position error as well as the external forces exerted at the end-effector. However, high forces are required to perturb the system when autonomous, as the stiffness is high in such a case.
Moreover, these approaches do not tackle energy considerations, which are paramount to ensure the system's stability.  
The stability of such schemes is often addressed by introducing energy tanks, 
that ensure the system passivity with time varying damping~\cite{kronander2015passive} or stiffness~\cite{ferraguti2015energy}.
Recent approaches~\cite{saveriano2020energy, michel2023passivity} show that the dynamics of the energy tank can be designed to further ensure stability in case of free motion.

In this work, we develop an incremental learning approach to an industrial human-robot collaborative application in which robot is required to execute a periodic task, and the worker can physically interact with the robot by updating its desired trajectory through kinesthetic teaching.
While the taught Cartesian trajectory is encoded with DMPs, the desired force is constant and specified by the task requirements. The demonstrated task's period is estimated through a custom adaptive frequency oscillator~\cite{petrivc2011line}.
The proposed strategy is able to interleave learning and execution phases without the need of two different controller strategies, by regulating the level of compliance, formulated as a quadratic program (QP) optimisation, according to the measured interaction, the tracking error, and the energy exchanged by the system. 
The energy tank not only ensures the system's passivity,
but is also used in the optimisation problem to regulate the input/output energy flow. By monitoring such flow, the system can distinguish desired demonstrations from other types of physical interactions, such as disturbances.  
We evaluate the proposed approach with a proof-of-concept polishing/cleaning/wiping task on arbitrary-shaped surfaces, where three tasks, with different positions and orientations, are successfully taught with few demonstrations and executed on-the-fly by the robot.

\section{Methodology} \label{sec:method}
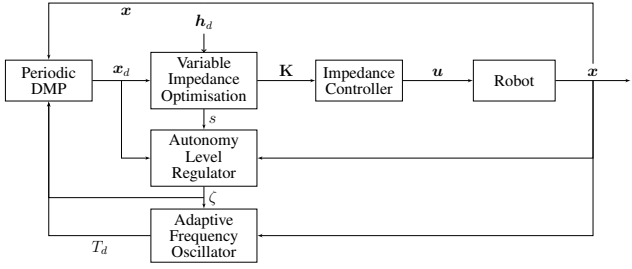
\begin{figure}[t!]
    \centering
    \adjustbox{width=1\columnwidth}{\tikzstyle{block} = [draw, fill=white, rectangle, 
    minimum height=3em, minimum width=6em]
\tikzstyle{sum} = [draw, fill=white, circle, node distance=1cm]
\tikzstyle{input} = [coordinate]
\tikzstyle{output} = [coordinate]
\tikzstyle{pinstyle} = [pin edge={to-,thin,black}]

\begin{tikzpicture}[auto, node distance=2cm,>=latex']

    \node [input, name=input] {};
    \node [block, right of=input, text width=2cm, align=center] (dmp) {\large Periodic DMP};
    \node [block, right of=dmp, node distance=4cm, text width=2.5cm, align=center, pin={[pinstyle]above:\large$\vec{h}_{d}$},] (vi_opt) {\large Variable Impedance Optimisation};
    \node [block, right of=vi_opt, node distance=4cm, text width=2cm, align=center] (imp_ctrl) {\large Impedance Controller};
    \node [block, right of=imp_ctrl, node distance=4cm, text width=1.5cm, align=center] (robot) {\large Robot};
    \node [output, right of=robot, node distance=3cm] (output) {};
    \node [block, below of=vi_opt, text width=2.5cm, align=center] (automation) {\large Autonomy Level Regulator};
    \node [block, below of=automation, text width=2.5cm, align=center] (osc) { \large Adaptive Frequency Oscillator};
    \coordinate [above of=vi_opt] (pos_feedback_loop) {};

    \draw [->] (dmp) -- node[name=x_d] {\large $\vec{x}_d$} (vi_opt);
    \draw [->] (vi_opt) -- node[name=k] {\large $\mat{K}$} (imp_ctrl);
    \draw [->] (imp_ctrl) -- node[name=u] {\large $\vec{u}$} (robot);
    \draw [->] (robot) -- node[name=x] {\large $\vec{x}$}(output);
    
    \draw [->] (x) |- (automation);
    \draw [->] (x_d) |- (automation);
    \draw [->] (vi_opt) -- node[name=tank] {\large $s$} (automation);

    \draw [->] (x) |- (osc);
    \draw [->] (automation) -- node[name=z] {\large $\zeta$} (osc);
    \draw [->] (z) -| (dmp);
    \draw [->] (osc) -| node[near start] {\large $T_d$} (dmp);

    \draw [-] (x) |- (pos_feedback_loop);
    \draw [->] (pos_feedback_loop) -| node[near start] {\large $\vec{x}$} (dmp);




\end{tikzpicture}}
    \caption{High level overview of the proposed framework.}
    \label{fig:framework}
    \vspace{-6mm}
\end{figure}

To this aim, we integrate three modules that work interactively to learn, execute, and adapt the required periodic motion. 
In particular, low-level control is relegated to a variable impedance controller that modulates the stiffness to generate a target force at the contact point. 
Interaction safety with the human demonstrator is ensured by energy tanks embedded in an optimisation problem that modulates the stiffness and avoids the instantaneous generation of excessive forces.
The end-effector trajectory is incrementally learned from a human demonstration and encoded into a periodic DMP that considers both position and orientation. To let the system work without any prior information regarding the demonstration period, we add to the framework an adaptive frequency oscillator which online estimates the fundamental frequency directly from the demonstration data.
The different modules are coordinated by an autonomy level regulator, a unit that's responsible for identifying the autonomy level $\gamma \in [0, 1]$ of the system that, in turn, tunes the learning factor $\zeta \in [0, 1]$ that's used in both the dynamic movement primitive and the adaptive frequency oscillator. The overall framework is depicted in Fig.~\ref{fig:framework}.

\subsection{Variable Impedance Control}\label{ssec:optimization}
To have a singularity-free description of the robot end-effector state, we consider the pose $\vec x = \big(\vec p^\top,\quat^\top)^\top \in \mathbb R^7$ as encoded by its position $\vec p\in \mathbb R^3$ and a unit quaternion $\quat \in \mathbb S^3 \subset \mathbb R^4$ to  represent orientation.

Let $\vec h^{ext} = \big({\vec f^{ext}}^\top, {\vec \tau^{ext}}^\top\big)^\top\in \mathbb R^6$ be the generalised force applied on the end-effector, with $\vec f^{ext},\vec \tau^{ext}\in \mathbb R^3$ respectively a force and a torque, then the objective of a Cartesian variable impedance controller is to render the robot closed-loop dynamics into a mass-spring-damper model in the form
\begin{equation}
\vec{h}^{ext} = \mat{\Lambda}\ddot{\Tilde{\vec{x}}} + \mat{D}\dot{\Tilde{\vec{x}}} + \mat{K}\Tilde{\vec{x}}.
    \label{eq:impedancemodel}
\end{equation}
In (\ref{eq:impedancemodel}), symmetric positive-definite matrices $\mat \Lambda,\mat D,\mat K \in \mathbb R^{6\times 6}$ are respectively the desired Cartesian inertia, damping and stiffness; to have a geometrically consistent definition of rotation difference in the quaternion domain, the Cartesian error is defined as 
\begin{equation}
    \Tilde{\vec x} = \begin{pmatrix}
        \Tilde{\vec p} \\ \Tilde{\quat}
    \end{pmatrix} =  \begin{pmatrix}
        \vec p^{des} - \vec  p \\
        \textrm{Log}(\quat^{des}* \overline \quat)
    \end{pmatrix}\in \mathbb R^6,
\end{equation}
with $\vec x_d = ({\vec p^{des}}^\top,{\quat^{des}}^\top)^\top$ the desired pose, $\textrm{Log}(\cdot): \mathbb S^3 \mapsto  \mathbb R^3$ the logarithmic map of the quaternion manifold, $*$ the quaternion multiplication operator, and $\overline \quat$ the conjugate quaternion of $\quat$.
To facilitate the analysis, (\ref{eq:impedancemodel}) is expressed in the compliance reference frame with respect to which the system is principal, i.e. $\mat \Lambda,\mat D,\mat K$ are all diagonal. In this paper, we chose as compliance frame the tool centre point.

Starting from (\ref{eq:impedancemodel}), we plan to module the desired stiffness $\mat K$ (and, accordingly, the damping $\mat D$) to generate some desired interaction behaviour depending both on the autonomy level and the task to accomplish.
In this scenario, it is known that a variable stiffness controller may generate non passive control actions~\cite{ferraguti2015energy}. Therefore, we enforce the passivity of the power port $\dot{\Tilde{\vec x}}^{\top}\boldsymbol{h}^{ext}$ using energy tanks~\cite{ferraguti2015energy}. To this end, we consider the tank dynamics
\begin{equation} \label{eq:tank_dynamics}
    \dot{{s}} = \frac{\sigma}{s}\dxtilde^\top \mat D \dxtilde - \frac{\vec k^\top}{s} \dxtilde,
\end{equation}
where ${s} \in \mathbb{R}$ is the state of the tank, $\sigma \in \{0,1\}$ is a coefficient modulating the energy storage, and $\boldsymbol{k}(t)$ is an extra input defined as:
\begin{equation} \label{eq:variable_stiffness}
    \vec k(t) = 
    \begin{cases} 
        -\mat{K}^\delta\Tilde{\boldsymbol{x}} & \textrm{if } T(s)>T_{min}  \\ 
        0 & \textrm{otherwise},
    \end{cases}
\end{equation}
with $T(s) = \frac 1 2 s^2$ the tank energy, which activates the extra input whenever is contains a minimum energy quantity $T_{min} \in \mathbb R^+$. 
In addition, let $\mat K^{min} \in \mathbb R^{6\times6}$ be the minimum allowed stiffness for the impedance controller, then $\mat K^{\delta} \in \mathbb R^{6\times6}$ is the time-varying component of the stiffness, thus satisfying $\mat K(t) = \mat K^{min} + \mat K^{\delta}(t) $. 

With~\eqref{eq:tank_dynamics} and~\eqref{eq:variable_stiffness}, to guarantee the system passivity, it is sufficient to choose a stiffness $\mat K$ that does not drive the energy tank below its minimum value $T_{min}$.

To modulate the stiffness of the impedance model~\eqref{eq:impedancemodel}, we cast the optimisation problem to a QP, thus ensuring that a solution can be found within each control period. 
In our framework, stiffness is modulated to accommodate two tasks: one aiming at producing a target wrench $\vec h^{des} \in \mathbb R^6$ at the end-effector, and a second regulating a desired stiffness $\mat K^{des} \in \mathbb R^{6\times 6}$~\cite{zhao2022hybrid}. 
The optimisation problem must take into account the energy tank constraint $T(s) \geq T_{min}$; this condition alone, however, does not prevent the energy tank from an instantaneous drain by selecting a high-enough stiffness, leading to a complete loss of performance~\cite{beber2023passive}. 
To overcome this limitation, we add a constraint to the power flow of the energy tank, limiting the energy extraction rate to aower bound $\rho \in \mathbb R^-$, i.e. $\dot T(s) \geq \rho$. 
The resulting QP is formulated as \vspace{-1.5mm}
\begin{align}
    & \underset{\mat K \in \mathbb{R}^{6 \times 6} }{\text{arg min}} \: \frac{1}{2} \left( \left\| \vec{h}^{ext} - \vec{h}^{des} \right\|_{\mat Q}^2 + \left\| \mat K - \mat K^{des} \right\|_{\mat R}^2 \right) \nonumber \\ 
    & \text{subject to   } \nonumber \\
    & \mat{K}^{min} \preccurlyeq \mat{K}^d \preccurlyeq \mat{K}^{max} \nonumber \\
    &\vec{h}^{min} \preccurlyeq \vec{h}^{ext} \preccurlyeq \vec{h}^{max} \label{eq:general_QP_formulation} \\
    &- \xtilde^\top \mat K \dxtilde \leq \sigma \dxtilde^\top \mat D \dxtilde - \xtilde^\top \mat K^{min} \dxtilde + \frac{ T_{t-1} - T_{min}}{\Delta t} \nonumber \\
    &- \ptilde^\top \mat K_{\vec p} \dptilde \leq \sigma \dptilde^\top \mat D_{\vec p} \dptilde - \ptilde^\top \mat K^{min}_{\vec p} \dptilde - \rho_{\vec p} \nonumber \\
    &- \qtilde^\top \mat K_{\quat} \dqtilde \leq \sigma \dqtilde^\top \mat D_{\quat} \dqtilde - \qtilde^\top \mat K^{min}_{\quat} \dqtilde - \rho_{\quat}, \nonumber \vspace{-1mm}
\end{align} 
where $\mat{Q}, \mat{R}$ $\in \mathbb{R}^{6\times 6}$ are diagonal positive definite weighting matrices, $\mat{K}^{min}, \mat{K}^{max} \in \mathbb{R}^{6\times 6}$ are diagonal matrices representing the minimum and maximum allowed stiffness, $\boldsymbol{h}^{ext} \in \mathbb{R}^{6}$ is the wrench of the impedance interaction model~\eqref{eq:impedancemodel},  and $\vec{h}^{max}, \vec{h}^{min} \in \mathbb{R}^{6}$ respectively the maximum and minimum wrench that the robot is allowed to exert. The symbol $\preccurlyeq$ represents the matrix inequality.
The third to last constraint avoids the complete drainage of the tank given the energy $T_{t-1}$ at the previous control cycle that takes $\Delta t$ to execute. The last equations instead constraint energy extraction rate; to take into account different scales of  both position and orientation error, the power constraint has been divided in 2 individual pieces; in light of this, $\mat K_{\vec p}, \mat K_{\quat} \in \mathbb R^{3\times3}$ are the stiffness sub-blocks of $\mat K$ associated respectively to the position and orientation error, and $\rho_{\vec p}, \rho_\quat \in \mathbb R^-$ are the power limits for both position and orientation. 

\subsection{Incremental Learning of Periodic Tasks}\label{ssec:pDMP}
To incrementally learn the trajectory of the end-effector, we exploit the Dynamic Movement Primitive (DMP) framework~\cite{saveriano2023dynamic}. Periodic DMP encodes motion trajectories in a spring-damper like system 
\begin{equation} \label{eq:singledmp}
\begin{cases}
    \dot z = \Omega\big( \alpha \big(\beta(g-y) - z\big) + f(\phi)\big) \\
    \dot y = \Omega z \\
    \dot \phi = \Omega,
\end{cases}
\end{equation}
with $\Omega$ the inverse of the task period, $g$ the motion goal, and $f(\phi)$ is a non-linear excitation term that must be learned from demonstration. Conventionally, $f(\phi)$ is chosen as the linear combination of $N$ Gaussian kernels, i.e. 
\begin{equation}
    f(\phi) = \frac{\sum_{i=1}^N \Psi_i(\phi) w_i}{\sum_{i=1}^N\Psi_i(\phi)}, 
\end{equation}
with 
\begin{equation}
    \Psi_i(\phi) = \exp\Big(h_i \big(\cos(\phi-c_i) -1\big) \Big).
\end{equation}
Given that a demonstration $y_d, \dot y_d, \ddot y_p$ provides a desired shape $f_d$ for the forcing term as
\begin{equation}
    f_d(t_j) = \frac{\ddot y_d(t_j)}{\Omega^2} - \alpha\left( \beta\big(g - y_d(t_j)\big) - \frac{\dot y_d(t_j)}{\Omega} \right), 
\end{equation}
it is possible to incrementally learn the weights $w_i$ using a recursive least-squares method with forgetting factor $\lambda$ as follows:
\begin{equation} \label{eq:dmpupdate}
\begin{aligned}
    w_i(t_{j+1}) & = w_i(t_j) + \zeta \Psi_i(\phi_i) P_i(t_{j+1}) e(t_j) \\
    e(t_j) & = f_d(t_j) - w_i(t_j) \\
    P_i(t_{j+1}) & = \frac 1 \lambda \left( P_i(t_j) - \frac{P_i^2(t_j)}{\frac \lambda {\Psi_i} + P_i(t_j)} \right), 
\end{aligned}
\end{equation}
where the system is initialised with $w_i(0) = 0, P_i(0) = 1$ for $i=1, \dots, N$. Note that in~\eqref{eq:dmpupdate}, the update is weighted also by the learning factor $\zeta$ that is specific to the framework we propose.

While~\eqref{eq:singledmp} is suitable to encode position variables, non-singular descriptions of orientation inherently come with constraints, which makes their learning non-trivial. To overcome this issue, we use the formulation proposed in~\cite{abu2021periodic} that exploits the Riemaniann property of the quaternion manifold to encode orientation in a DMP:
\begin{equation} \label{eq:quaterniondmp}
\begin{cases}
    \dot{ \vec \eta} &= \mat\Omega \left(\alpha_z(\beta_z2 \, \mathrm{Log}(\quat*\overline{\quat}_d)-\vec{\eta}) + \vec{f}(\vec{\phi})\right) \\
    \dot{\quat} &= \mat{\Omega}\frac{1}{2}\vec{\eta}*\quat,
\end{cases}
\end{equation}
with $\mat \Omega = \Omega \mat I_{3\times 3}$. Differently from~\cite{abu2021periodic}, quaternion periodic DMP learning is performed incrementally using~\eqref{eq:dmpupdate} for each one of the 3 coefficients of $\vec f(\phi) \in \mathbb R^3$, thus no coupling in the forcing terms is considered.




\subsection{Online Period Estimation}\label{ssec:oscillator}
A crucial aspect in learning periodic motions is the estimation of the period of the demonstrated trajectory. To address this issue, we augment the adaptive frequency oscillator (AFO) proposed in~\cite{petrivc2011line} with the learning rate $\zeta$.
AOFs are dynamical systems whose goal is to learn the Fourier series,
truncated at the first order, of the input signal, which is, in this case, one component of the desired trajectory. Let $\hat l = \alpha \varphi \omega + \beta \cos \varphi \in \mathbb R $ be the first-order expansion of the signal $l \in \mathbb R$, and calling $e = l - \hat l \in \mathbb R$ the corresponding error, then parameters $\alpha,\beta,\omega \in \mathbb R$ are estimated integrating the dynamics: \vspace{-2mm}
\begin{equation} \label{eq:oscillator}
\begin{cases} 
    \dot \varphi = \omega - \zeta  a e \sin \varphi \\
    \dot \omega = -\zeta a e \sin \varphi \\
    \dot \alpha = \zeta \eta e \cos \varphi \\
    \dot \beta = \zeta \eta e \sin \varphi
\end{cases}.
\end{equation}
Here, $\omega, \varphi \in \mathbb R$ are respectively the basic frequency and the phase of the signal, while $\alpha,\beta$ are parameters modulating the amplitude of the reconstructed signal; $a \in \mathbb R^+$ is a coupling constant that drives the convergence of the frequency estimates, while $\eta \in \mathbb R^+$ is the learning coefficient of the modulating terms.
Once the dynamics of the AFO converges to the fundamental angular frequency $\omega$ of the demonstration, the corresponding period $T\in \mathbb R$ can be simply obtained as \vspace{-1mm}
\begin{equation} \label{eq:period}
    T = \frac{2\pi}{\omega}.
\end{equation}

In our case, multiple signal sources can be used to estimate the period of the motion, i.e. 3 for the Cartesian position and 4 for the quaternion trajectory of the end-effector. Even though each DMP is allowed to evolve at its own time constant $\tau$, we stick to a common period to avoid synchronisation issues between different tasks. To properly select the shared demonstration period $T_d$, we let each independent $i$-th source estimate its period $T_i$ according to (\ref{eq:oscillator}) and (\ref{eq:period}); then, adapting from~\cite{Dimes2020Progressive}, we select $T_d$ as \vspace{-3mm}
\begin{equation} \label{eq:periodselector}
    T_d = \arg \max_i \left\{T_i \frac{\overline l_i}{\max_j\{\overline l_j\}}\right\},
\end{equation}
where $\overline l_i = S\big(\textrm{Var}\{l_i\}\big) \in (0,1)$ is the logistic sigmoid function computed on the variance $\textrm{Var}\{l_i\}$ of the $i$-th signal. Tuning the steepness $a_l \in \mathbb R^+$ and the threshold $\sigma^2_0 \in \mathbb R^+$ of \vspace{-1mm}
\begin{equation} \label{eq:sigmoid}
    S(x) = \frac{1}{1 + e^{-a_l(x-\sigma^2_0)}},
\end{equation}
it is possible to effectively penalise period estimates associated to signals with low variance. For instance, while demonstrating a motion on the $x,y$ plane, the AFO associated with the $z$ coordinate, learning from noisy data, could converge to an unreasonable period. The scaling term $\overline l_i/\max\{\overline l_i\}$ in~\eqref{eq:periodselector} handles such conditions by artificially reducing the periods associated with less interesting signals, effectively discouraging their selection for the overall learning framework.

\subsection{Autonomy Level Regulator}\label{ssec:learning_rate}
We introduced an autonomy factor $\gamma \in [0,1]$ which is a variable describing the ability of the system to autonomously execute the task ($\gamma = 1$) or to stand still waiting for a new demonstration ($\gamma = 0$).
To this extent, we propose a modification of the automation dynamic proposed in \cite{Dimes2020Progressive} that embeds the energy tank state in order to relax requirements on the tracking error. In particular, we select the autonomy factor $\gamma(t)$ dynamics as described by
\begin{equation}
    \dot \gamma = \begin{cases}
        \max\{\gamma_r, 0\} \quad & \gamma = 0 \\
        \gamma_r & 0 < \gamma < 1 \\
        \min\{\gamma_r, 0\} & \gamma = 1
    \end{cases}, 
\end{equation}
with \vspace{-3mm}
\begin{dmath} \label{eq:autonomyrate}
    \gamma_r = S\big(\|\dot {\vec x}\|_{\mat G}^2\big) \big(\gamma_{r,min} + k_r \gamma\big) \\ \Big(1 - \nu_1 \|\xtilde\|_{\mat L}^2 - \nu_2 \max \{0, \xi - \xi_{min}\} \Big).
\end{dmath}
Here, $\gamma_{r, min}, k_r, \nu_1,\nu_2, \xi_{min} \in \mathbb R^+$ are all parameters that must be tuned, while $S(\cdot)$ is a sigmoid function, as in~\eqref{eq:sigmoid}, that is used to avoid premature learning while standing still.

In~\eqref{eq:autonomyrate}, we consider an additional state $\xi$ whose dynamic is associated to the power-flow of the energy tank according to the linear system \vspace{-1mm}
\begin{equation} \label{eq:lowpass}
    \dot \xi = \frac{1}{\tau_\xi} \Big( \dot T(s) - \xi\Big).
\end{equation}
b
With such choice, we effectively create a low-pass filter for the tank power signal, thus rejecting high-frequency disturbances that might arise during contact with rough surfaces. This, in addition with the threshold in~\eqref{eq:autonomyrate}, allows to quickly reset the autonomy factor whenever lots of energy is pumped in the tank, i.e., when the human is acting on the robot.



\section{Experiments} \label{sec:experimental}
\subsection{Setup}
\begin{figure}
    \centering
    \import{images/}{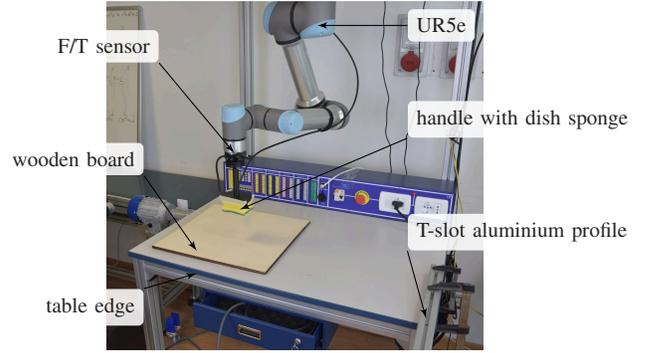}
    \vspace{-4mm}
    \caption{Experimental setup consisting of a ceil-mounted UR5e manipulator, the SenONE BOTA System force-torque sensor, the 3D printed handle and the dish sponge. For the experiment, there is a floating wooden board and a table-fixed T-slot aluminium profile.}
    \label{fig:setup}
    \vspace{-6mm}
\end{figure}
The framework is evaluated in a proof-of-concept robotic setup, depicted in Fig. \ref{fig:setup}. 
The system is made of a ceil-mounted 6 DoF manipulator, the Universal Robot UR5e, with force-torque sensor, the SensONE BOTA System, mounted between the wrist and the end-effector. 
For the demonstration, the tool is made by a 3D-printed handle, at which end is attached a dish sponge used for the polishing operation.

The UR5e robot is position controlled, thus direct compliance control is not possible. 
To overcome this technological limitation, we employ the Forward Dynamics Compliance Control (FDCC) algorithm~\cite{scherzinger2017forward}, a method that enables impedance, admittance, and force control in a single strategy for position-controlled robots. 


To evaluate the proposed framework, we test the system while performing a wiping task on surfaces of different materials.
To achieve the goal, we set as desired wrench for the QP problem~\eqref{eq:general_QP_formulation} the one with only a positive action oriented along the positive $z$ axis of the end-effector:
\begin{equation}
    \vec h^{des} = \big(0, 0, 5\textrm{N}, 0, 0, 0\big)^\top.
\end{equation}
To have a robot which is fully compliant while learning ($\gamma = 0$), and stiff when autonomous ($\gamma = 1$) to reduce tracking errors, the desired stiffness is modulated as
\begin{equation} \label{eq:Kdes}
    \mat K^{des} = \mat K^{min} + \gamma^2 \big( \mat K^{am} - \mat K^{min}\big),
\end{equation}
with $\mat K^{min} = \blkdiag{\mat I_{3\times 3}\,\textrm{N/m}, 0.001 \, \mat I_{3\times 3}\,\textrm{Nm/rad}}$ and $\mat K^{am} = \diag{600, 600, 600, 60, 60, 70}$ the stiffness in autonomous mode; for the experiment, maximum allowed stiffness has been set to $\mat K^{max} = \blkdiag{1000\mat I_{3\times 3}, 100\mat I_{3\times 3}}$.

To have a proper trade-off between the two complementary tasks in (\ref{eq:general_QP_formulation}), tuning the weighting matrices $\mat Q, \mat R$ is of key importance. 
In this case, we set the former to $\mat Q = \blkdiag{3200\mat I_{3\times 3}, 32\mat I_{3\times 3}}$, while the latter is modulated, similarly to~\eqref{eq:Kdes}, as $\mat R = \mat R^{min} + \gamma^2 (\mat R^{am}-\mat R^{min})$, with $\mat R^{min} = \blkdiag{0.1 \mat I_{3\times 3}, 0.01\mat I_{3\times 3}}$ and $\mat R^{am} = \diag{10, 10, 1.0,  10, 10, 10}$.
These choices are made to guaranteed maximum compliance for $\gamma = 0$, while in autonomous mode the controller can freely modulate the stiffness to track the force on the $z$ component, while translation in the $x,y$  plane and orientation is regulated through bigger stiffness gains.
To complete the QP problem definition, we set $\vec h^{max} = (40, 40, 40, 10, 10, 10)$, $\vec h^{min} = - \vec h^{max}$, $T_{min} = 0.1$, $\rho_{\vec p} = -1.1$, and $\rho_\quat = -1,4$.

Concerning the DMPs, we set $\alpha = 48$, $\beta = 12$, $N=25$ basis functions and forgetting factor $\lambda = 0.999$ for both position and orientation DMP, while their goals are initialised to the starting configuration of the robot (to avoid unexpected movements when the robot starts with no excitation term learned).
Gains of the adaptive frequency oscillator are set to $a = 50, \eta = 1$, and to avoid divergence of the dynamical system, we constrained $\omega \in [0.7, \pi]$ and $|\alpha| = |\beta| \leq 0.3$.

Finally, the parameters for the autonomy level dynamics are $\gamma_{r,min} = 0.1$, $k_r = 0.25$, $\nu_1 = 450$, $\nu_2 = 15$, $\xi_{min} = 0.8$, $\tau_\xi = 0.5$, $\mat L = \diag{1, 1, 0.2, 0.1, 0.1, 0.1}$ and $\mat G = \blkdiag{\mat I_{3\times 3}, 0.1\mat I_{3\times 3}}$. In the following demonstration, the learning rate $\zeta$ is computed as $(1-\gamma)^2$.

\subsection{Results and Discussion}
\begin{figure*}[t!]
    \centering
    \def\imgwidth{2.2cm}
    \begin{tabular}{c c c c c c c c}
        \includegraphics[width=\imgwidth, angle=90]{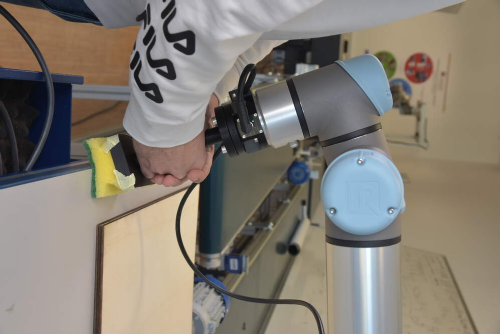} &
        \includegraphics[width=\imgwidth, angle=90]{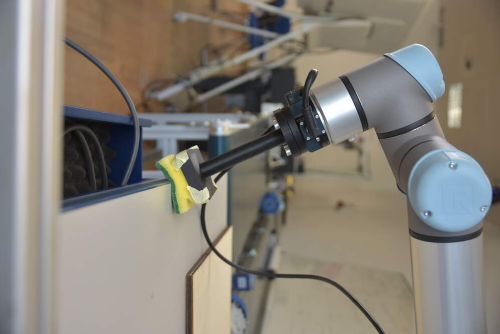} &
        \includegraphics[width=\imgwidth, angle=90]{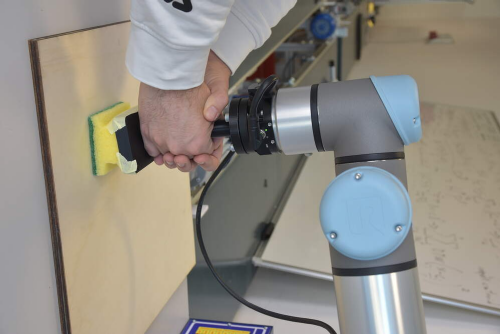} &
        \includegraphics[width=\imgwidth, angle=90]{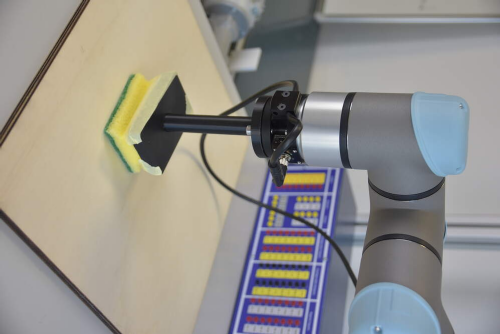} &
        \includegraphics[width=\imgwidth, angle=90]{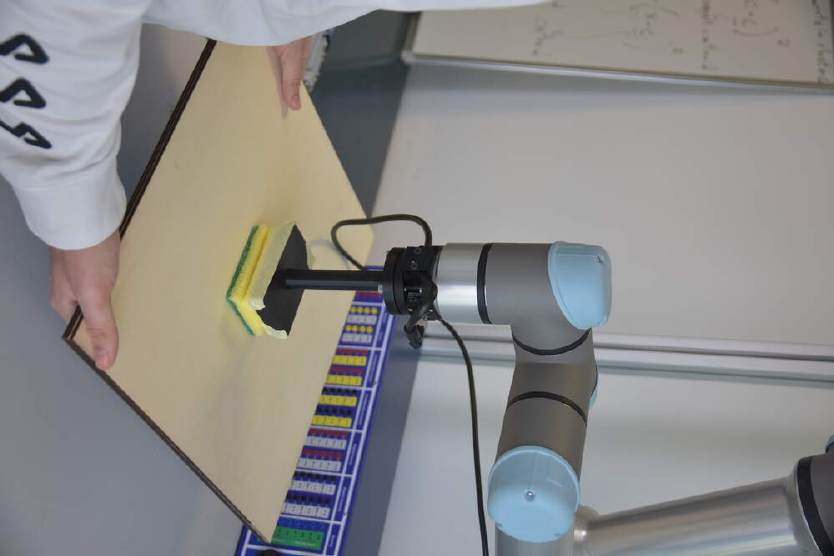} &
        \includegraphics[width=\imgwidth, angle=90]{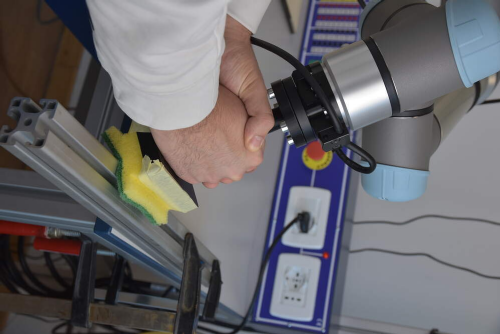} &
        \includegraphics[width=\imgwidth, angle=90]{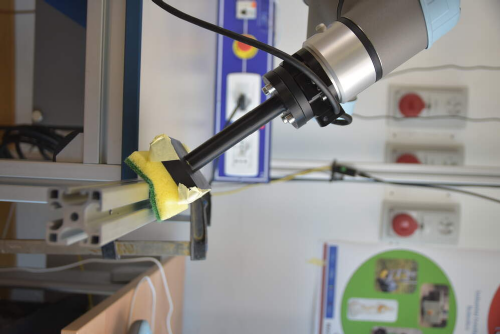} &
        \includegraphics[width=\imgwidth, angle=90]{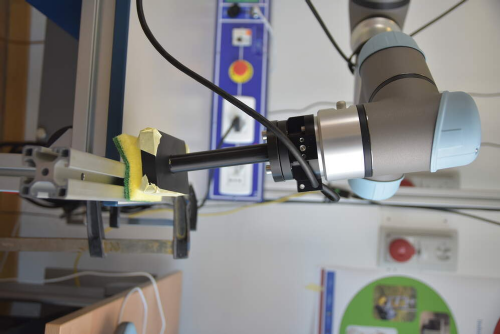} 
         \\
         \footnotesize (a) & \footnotesize (b) & \footnotesize (c) & \footnotesize (d) & \footnotesize (e) & \footnotesize (f) & \footnotesize (g)  & \footnotesize (h)
    \end{tabular}
    \includegraphics[width=\linewidth]{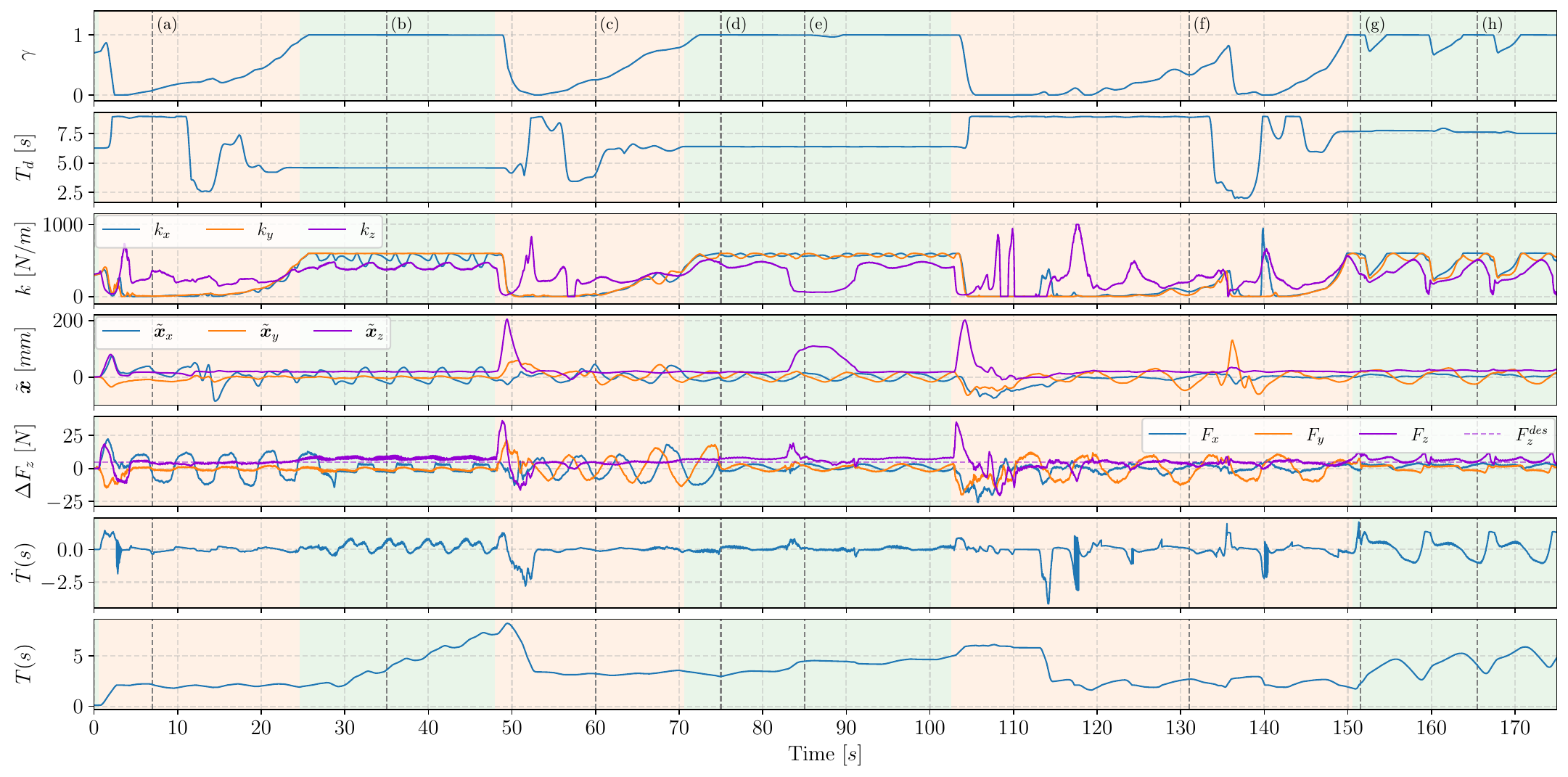} \vspace{-8mm} 
    
    \caption{Snapshots taken from the demonstration (available at \href{https://youtu.be/2fCSnqNbqz0}{https://youtu.be/2fCSnqNbqz0}) and collected data: (a) the human demonstrates how to clean the edge of a table, and (b) the robot executes it. Then (c), the human demonstrates a circular motion on a wooden table, and (d) the robot executes it (e), even when the object is perturbed. Finally, (f) the human shows a motion that requires the tilting of the robot end-effector, as for the poses shown in (g) and (h). 
    Green areas represent time windows where the robot is autonomous, while orange ones represent time windows when the human is interacting.
    }
    \label{fig:demo}
    \vspace{-4mm}
\end{figure*}

To show the effectiveness of the proposed framework, we run experiments involving the teaching of different wiping tasks within the same robot execution, proving how the system can easily adapt to different requirements. In particular, Fig.~\ref{fig:demo} reports a demonstration in which the robot: 1) starts from  a standstill configuration, then 2) the operator demonstrates how to clean the edge of the table by moving along it and after a few autonomous executions, 3) the system is taught a circular motion on a wooden surface. 
At this stage, 4) we show the safety of the proposed optimised variable impedance controller by moving the plane on which the robot is executing his wiping task. Finally, 5) the human operator demonstrates how to polish a T-slot aluminium profile with a motion that requires the end-effector re-orientation to properly carry out the task.

At the beginning of the demonstration, when the human takes over the robot, energy is injected in the tank and the autonomy factor drops to $0$.
One can observe a time delay between the proper energy injection and the factor drop, but this is due to the low-pass filter~\eqref{eq:lowpass}.
Additionally, while learning from a new demonstration, the automation sometimes dips down: this is related to the adaptive frequency oscillator that, at the same moment, converges to the actual demonstration period. 
This sudden change in the estimated period negatively affects the learning performance of the DMP, but convergence at the desired motion is restored quickly. 
To properly converge at the desired motion with no prior knowledge of the task, the system takes about $20s$ for planar tasks, i.e. 4 to 5 full demonstrations.
In case of a demonstration involving the active learning of an orientation trajectory, the convergence rate is slower due to two primary reasons: firstly, the convergence rate of the orientation DMP is slower, and, secondly, the virtualisation mechanism that enables compliance control of a position-controlled robot generates an apparent rotational stiffness, increasing the difficulty of humans to demonstrate repeatable motion patterns. 
However, the system is able to converge even slowly in this scenario. Additionally, these issues can be mitigated using torque-controlled robots for which compliance control is properly defined, without the need for virtualisation mechanisms.

During autonomous execution, the optimisation problem modulates the stiffness in the $x, y$ plane towards the $\mat K^{am}$ value, leading to a tracking error within the $2mm$ set for the application; to reduce tracking error, it is sufficient to increase $\mat K^{am}$. 
Concerning the force-tracking requirements along the $z$ axis, actual applied forces stay mainly within the $5\,$N error range. If one seeks better force regulation, $\mat R^{am}$ can be accordingly adjusted. Still, during execution, we preferred a trade-off toward a higher stiffness to guarantee more contact in situations at which the DMP provides targets just slightly below the surface. 
Finally, the choice~\eqref{eq:Kdes} of the desired stiffness is a valuable feedback for the demonstrator, since as the system gets autonomous, the stiffness is perceived back by the human.

\section{Conclusion} \label{sec:conclusions}
In this paper, we present our preliminary work on a incremental learning framework that combines a safe variable impedance control, adaptive frequency oscillator, and dynamic movement primitives to perform wiping tasks.
Differently from previous works in the field, the DMP also learns the orientation, thus, we extended the problem formulation of variable stiffness optimisation to the orientation. 
In doing so, we enabled the execution of the wiping task on any smooth surface, not necessarily flat ones.
In addition, we propose a novel formulation for autonomy-level dynamics that embeds the energy-tank knowledge to relax tracking error satisfaction to drive the learning phase of the module, while, in the meantime, guaranteeing a good disturbance rejection.
In future works, we will address some limitations of the proposed architecture. First, we will join the dynamic movement primitive module with the adaptive frequency oscillator to enable a faster convergence of the DMP towards the demonstrated task, even in conditions where end-effector re-orientation is required. 
Second, we will integrate continual learning techniques to avoid the inefficient loss of the acquired demonstration knowledge that, as of now, is discarded by the system. 

\balance 
\bibliographystyle{IEEEtran}
\bibliography{ref}

\end{document}